\documentclass[10pt,twocolumn,letterpaper]{article}
\usepackage[accsupp]{axessibility}
\usepackage{cvpr}              

\usepackage{graphicx}
\usepackage{amsmath}
\usepackage{amssymb}
\usepackage{booktabs}
\usepackage[flushleft]{threeparttable}

\usepackage[pagebackref,breaklinks,colorlinks]{hyperref}

\usepackage[capitalize]{cleveref}
\crefname{section}{Sec.}{Secs.}
\Crefname{section}{Section}{Sections}
\Crefname{table}{Table}{Tables}
\crefname{table}{Tab.}{Tabs.}

\def\cvprPaperID{10864} 
\def\confName{CVPR}
\def\confYear{2022}

\begin{document}

\title{Searching the Deployable Convolution Neural Networks for GPUs}

\author{Linnan Wang, Chenhan Yu, Satish Salian, Slawomir Kierat, Szymon Migacz  \\ Alex Fit Florea\\
NVIDIA\\
}
\maketitle

\begin{abstract}
Customizing Convolution Neural Networks (CNN) for production use has been a challenging task for DL practitioners. This paper intends to expedite the model customization with a model hub that contains the optimized models tiered by their inference latency using Neural Architecture Search (NAS). To achieve this goal, we build a distributed NAS system to search on a novel search space that consists of prominent factors to impact latency and accuracy. Since we target GPU, we name the NAS optimized models as GPUNet, which establishes a new SOTA Pareto frontier in inference latency and accuracy. Within 1$ms$, GPUNet is 2x faster than EfficientNet-X and FBNetV3 with even better accuracy. We also validate GPUNet on detection tasks, and GPUNet consistently outperforms EfficientNet-X and FBNetV3 on COCO detection tasks in both latency and accuracy. All of these data validate that our NAS system is effective and generic to handle different design tasks. With this NAS system, we expand GPUNet to cover a wide range of latency targets such that DL practitioners can deploy our models directly in different scenarios.
\end{abstract}

\section{Introduction}
\label{sec:intro}

The progress of neural networks has decoupled from the actual deployment for a long time. Deep Learning (DL) researchers have been dedicated to inventing new building blocks, while DL engineers deploy these building blocks in real-world tasks, painstakingly recombine them to find architectures that meet the design requirements. Most of the time, we can simplify these requirements to find the best-performing architecture on the target device (e.g., GPUs) within a specific latency budget. Though there are many exciting advancements in the neural network designs, e.g., the residual connection~\cite{he2016deep}, Inverted Residual Block (IRB)~\cite{sandler2018mobilenetv2} and the attention~\cite{vaswani2017attention}, deploying these network designs remains challenging and laborious; and this is the problem to be addressed in this paper.

Our solution to alleviate the gap between the DL research and the actual deployment is to propose a set of optimized Convolution Neural Networks for each type of GPUs tiered by their optimized inference latency (e.g., post-processed by TensorRT~\cite{TensorRTIntro} or OpenVINO~\cite{openVINOIntro}). Specifically, we deliver a table of models, an entry of which is the result of model optimization from maximizing the accuracy subject to the limit of inference latency on a GPU. This table enables DL engineers to directly query the optimized neural architecture w.r.t the design requirements to expedite the customization process on expensive models.

\begin{figure}[t]
\centering 
  \begin{center}
    \includegraphics[width=0.95\columnwidth]{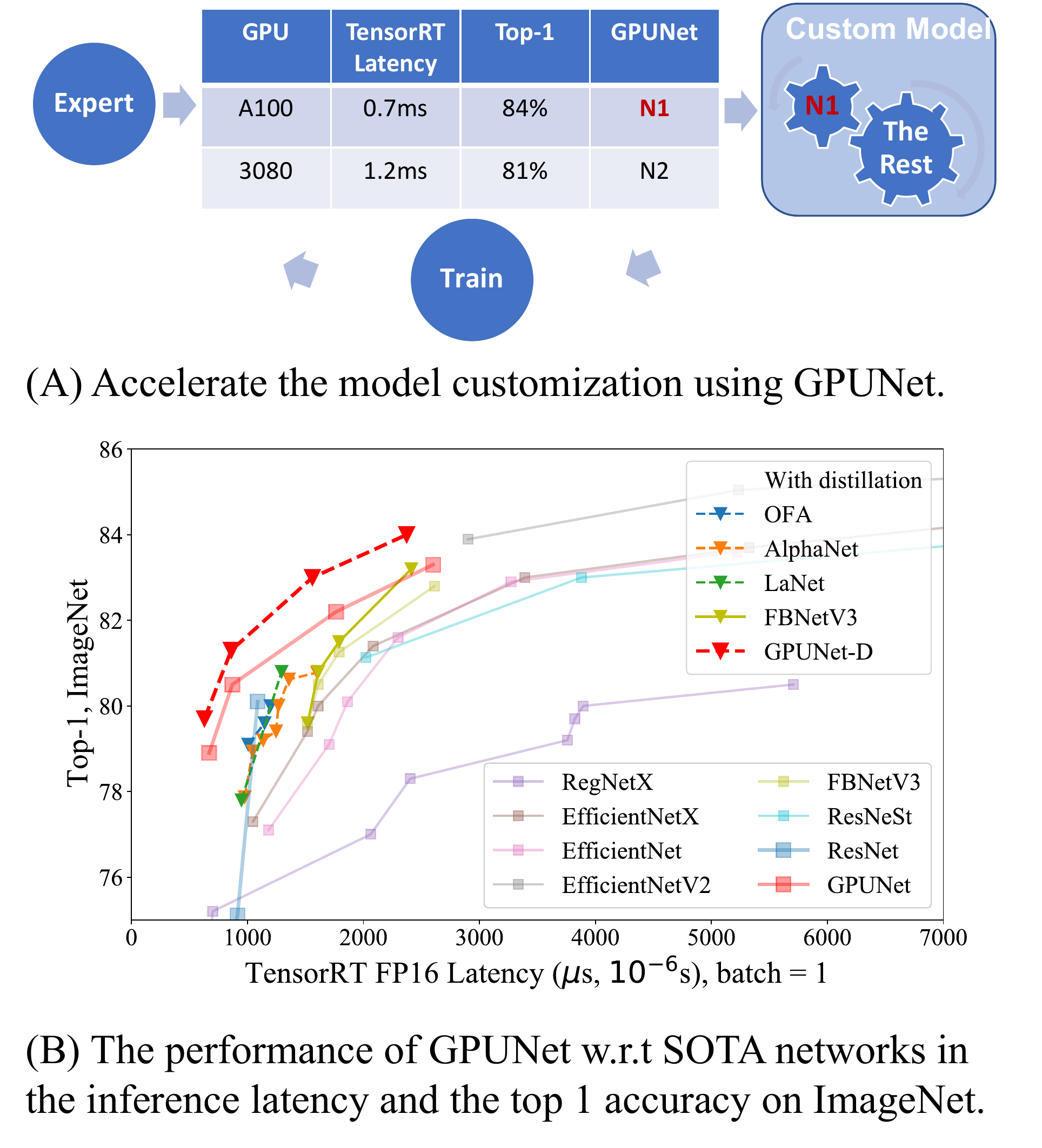}
  \end{center}
     \vspace{-0.5cm}
    \caption{GPUNet establishes the new SOTA Pareto frontier in the accuracy and inference latency.}
    \label{teaser_figure}
    \vspace{-0.2cm}
\end{figure}

We resort to Neural Architecture Search (NAS) to design models in this table. Recently NAS has shown promising results to automate the design of network architectures in many tasks~\cite{wang2019alphax, wang2021sample, liu2019auto}. Therefore, NAS can be a handy tool since we need to design many models for many latency limits for different GPUs. When models are ready to deploy, we measure post-processed TensorRT engine latency, i.e., including quantization, layer/tensor fusion, kernel tuning, and other system side model optimizations. Finally, we design our model toward NVIDIA enterprise GPU products for their broad adoption by the community today.

We built a novel distributed NAS system to achieve our goal. Following the prior works, our NAS consists of 3 modules, a search space, an evaluation module, and a search method. The search space provisions networks following the predefined patterns; the search method proposes the most promising network based on the priors. The evaluation module returns the performance of the proposed network either by training or estimation from a supernet~\cite{zhao2021few}. Our search space constructs a network by stacking convolution layers, IRBs, and Fused-IRBs used in EfficientNet~\cite{tan2021efficientnetv2}. However, our search space is the most comprehensive by far that includes filter numbers (\#filters), kernel sizes, the number of layers (\#layers) or IRBs (\#IRBs) in a stage, and the input resolution. Within an IRB or Fused-IRB, we also search for the expansion ratio, the activation type, with or without the Squeeze-Excitation (SE) layer. All of these factors are identified as prominent factors to affect latency and accuracy. Therefore, this search space enables us to better leverage the accuracy and latency than prior works, e.g., the fixed filter pattern in NASNet~\cite{zoph2018learning} and the fixed activation and SE pattern in FBNetV3~\cite{dai2021fbnetv3}; and the search also enables us to find a better policy than the fixed scaling strategy in EfficientNet~\cite{li2021searching, tan2019efficientnet}. To support such a complex search space, we choose to evaluate a network candidate by training. Although this approach is far more expensive than the supernet approach, the evaluation is more accurate in ranking the architectures~\cite{yu2019evaluating, zhao2021few}. And we can avoid many unresolved problems in building a supernet for our search space, e.g., supporting multiple types of activation, activating/deactivating SE, and variable filter sizes. We built a client-server-style distributed system to tackle the computation challenges, and it has robustly scaled to 300 A100 GPUs (40 DGX-A100 nodes) in our experiments. Finally, we adopt the LA-MCTS guided Bayesian Optimization (BO)~\cite{wang2020learning} as the search method for its superior sample efficiency demonstrated in the recent black-box optimization challenges~\cite{BBOC}.

We name the NAS optimized CNNs as GPUNet, and GPUNet has established a new SOTA Pareto front in the latency and accuracy in Fig.~\ref{teaser_figure}. We measure the latency of GPUNet using TensorRT, so GPUNet is directly reusable to DL practitioners. Particularly, GPUNet-1 is nearly 2x faster and 0.5\% better in accuracy than FBNetV3-B and EfficientNet-X-B2-GPU, respectively. We also validate GPUNet on COCO detection tasks, and GPUNet still consistently outperforms EfficientNet-X and FBNetV3. All of these data validate that our NAS system is effective and generic in designing various tasks. Although this paper only shows a few GPUNet for comparisons, the complete model hub tiered by the inference latency is still ongoing, and we will release them with the paper.

\section{Related Works}

Today running DL models locally on an edge device or hosting the model as a service on the enterprise-level GPUs in data centers are two major ways for the model deployment. Edge devices, such as your smartphones or the embedded system for the self-driving, are often equipped with a small CPU, limited RAM, and slow inter-connect. Given the constrained resource, a model's \#{\tt FLOPS} or MACs can correlate well with the latency on such devices, driving many works to propose low {\tt FLOPS} operators for the faster deployment on the edge devices~\cite{howard2017mobilenets, sandler2018mobilenetv2, ramachandran2017searching}, e.g., depth-wise separable convolutions. Accelerating the inference on mobile devices has been a hot research topic in recent years~\cite{cai2018proxylessnas, wu2019fbnet, howard2019searching}.

Deploying models on the GPU requires different optimizations from the edge devices. Since GPU has a far more powerful architecture than edge devices, we need to consider the device saturation, parallelism and compute/memory-bound operators, etc., for the deployment~\cite{li2021searching}. Although GPU has dominated the MLPerf inference benchmark for years, only a few works optimize the CNN deployment on GPUs. One line of these works is to propose fast operators. RegNet~\cite{radosavovic2020designing} accelerates the inference by optimizing the search space to select simple, GPU-friendly operators. ResNetXt proposes the split attention operator to improve the accuracy and inference latency~\cite{zhang2020resnest}. Another line of work is to optimize the structure. TResNet~\cite{ridnik2021tresnet} optimizes operators in ResNet-50, including SE layers and BatchNorm, to improve the inference on GPU. At the same time, EfficientNet-X~\cite{li2021searching} proposes a latency-aware scaling method for designing the fast EfficientNet to the GPU/TPU and uses a roofline model to explain the gap between the {\tt FLOPS} and latency on GPUs. Rather than using a fixed scaling policy, our work treats the model optimization as a black box, searching for the fast architectures for GPUs using the TensorRT optimized inference latency. Therefore, we can better trade-off the latency and accuracy than EfficientNet-X, and our final networks are directly deployable on GPUs.

We use NAS to build proposed networks, and here we review the recent advances in NAS. Early works in NAS, e.g., NASNet~\cite{zoph2018learning}, models CNN as a Direct Acycle Graph (DAG). While EfficientNet quickly gains popularity for its good performance on ImageNet~\cite{xie2020self}. This paper reuses the building blocks from EfficientNetV2 to find fast architectures. Recently transformer~\cite{liu2021swin} and Multi-Layer Perception (MLP)~\cite{liu2021pay} starts to emerge as promising alternatives to ConvNet; we leave the NAS on these search spaces as future work. Our paper evaluates each network independently by training end-to-end despite the popularity of the supernet approach. ENAS~\cite{pham2018efficient} proposed the supernet, which is an over-parameterized network to approximate the performance of sub-networks. Although the supernet significantly reduces the computation requirement for NAS, the rank predicted by supernet can be inaccurate~\cite{zhao2021few}. Besides, training the supernet is non-trivial~\cite{yu2020train} and the construction of supernet to support variable activation, expansion ratio, and filter sizes, etc., still remains an open problem. Therefore, we perform NAS using a distributed system to avoid unresolved issues of supernet.

\begin{figure*}[t]
\vspace{-0.5cm}
\centering 
  \begin{center}
    \includegraphics[width=0.97\textwidth]{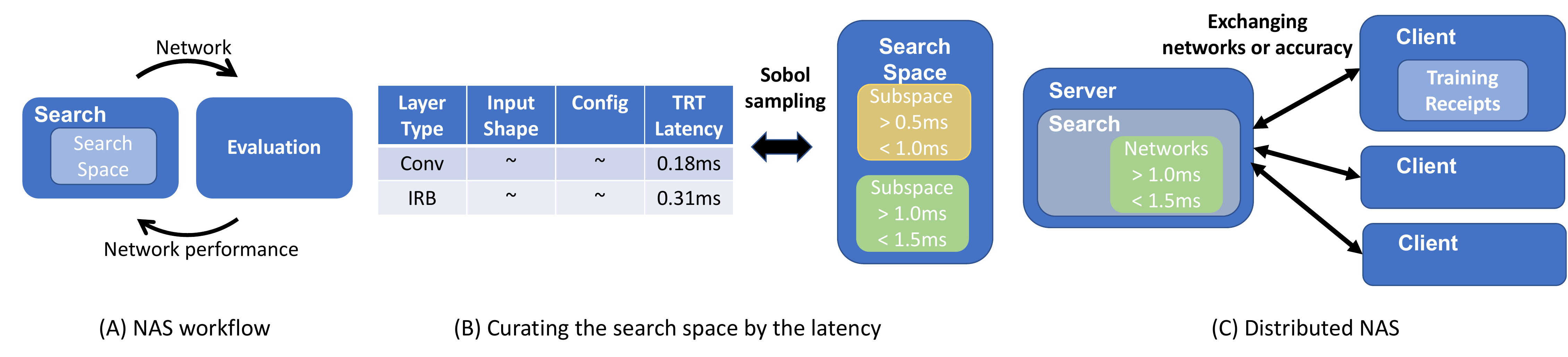}
  \end{center}
      \vspace{-0.5cm}
    \caption{The work flow of proposed NAS framework. A proposed search space is first pruned by TensorRT inference latency in (B). Then we use a black box optimizer to iteratively explore the search space in (A). We implement
    a distributed search framework to exploit the parallelism in (C).}
    \label{fig:nas_framework}
\end{figure*}
\section{Methodology}

We built a novel distributed NAS framework to automate the model design for GPUs. Our NAS system consists of a search space, a search algorithm, and an evaluation method following the existing NAS framework. First, the search algorithm selects networks from the search space, querying its performance using the evaluation method. Then the search algorithm refines its decision in the next iteration by leveraging all the evaluated network accuracy pairs. 

Our NAS consists of two stages, 1) categorizing networks by the inference latency and 2) performing NAS on networks within a latency group to optimize the accuracy. In the first stage (Fig.~\ref{fig:nas_framework}.A), we use Sobol sampling~\cite{sobol1967distribution} to draw network candidates from the high-dimensional search space evenly, approximate the network latency by using the latency look-up table, then categorize the network into a sub-search space, e.g., networks $<$ 0.5ms. We approximate the inference latency by summing up the latency of each layer from a latency lookup table. The latency table uses the input data shape and layer configurations as the key to the latency of a layer. In the second stage (Fig.~\ref{fig:nas_framework}.B), Bayesian optimization consumes a sub-space to find the best performing network within the latency range of the sub-space. We built a client-server distributed framework to perform NAS. The search algorithm runs on the server, proposing network for a client. The client will return the accuracy and network after training. The following elaborates each components and its design justifications.

\subsection{Search Space}
\label{sec:search_space}

The search space prescribes the general structure of network candidates, and our search space is inspired by EfficientNet~\cite{tan2021efficientnetv2}. Please note our search framework is generic to support various search spaces, e.g., designing visual transformer or MLP for visual tasks. Although transformer has shown excellent performance recently~\cite{liu2021swin, bao2021beit}, here we focus on ConvNet due to the better support from current TensorRT that performs critical performance optimizations for the fast inference on GPUs. We leave the NAS on the transformer- or MLP-based vision models as future work.

\begin{table}[!tb]
\begin{threeparttable}
\footnotesize
\setlength{\tabcolsep}{0.2em}
  \centering
    \begin{tabular}{ l l l l l l l l l}
          \toprule
          {Stage} & {Type} & Stride & Kernel & \#Layers        & Act   & E$^{\ddagger}$ & Filters & SE   \\
    	  \midrule 
    	  0       & Conv       & 2      & [3, 5] & 1         & [R,S]$^\mathbf{\dagger}$ &       & [24, 32,  8]$^*$ &        \\
		  1       & Conv       & 1      & [3, 5] & [1, 4]    & [R,S]           &       & [24, 32,  8]     &        \\
		  \midrule
		  2       & F-IRB$^\diamond$  & 2      & [3, 5] & [1, 8]    & [R,S] & [2, 6] & [32, 80,  16]    & [0, 1] \\
		  3       & F-IRB             & 2      & [3, 5] & [1, 8]    & [R,S] & [2, 6] & [48, 112, 16]    & [0, 1] \\
		  \midrule
		  4       & IRB        & 2      & [3, 5] & [1, 10]   & [R,S] & [2, 6] & [96, 192, 16]    & [0, 1] \\
		  5       & IRB        & 1      & [3, 5] & [0, 15]   & [R,S] & [2, 6] & [112, 224, 16]   & [0, 1] \\
		  6       & IRB        & 2      & [3, 5] & [1, 15]   & [R,S] & [2, 6] & [128, 416, 32]   & [0, 1] \\
		  7       & IRB        & 1      & [3, 5] & [0, 15]   & [R,S] & [2, 6] & [256, 832, 64]   & [0, 1] \\
		  \midrule
		  8       & \multicolumn{6}{c}{Conv1x1 \& Pooling \& FC} & 1792 \\
	      \midrule
	      & Res & \multicolumn{6}{c}{[224, 256, 288, 320, 352, 384, 416, 448, 480, 512]} \\
	      \bottomrule
    \end{tabular}
    \begin{tablenotes}
      \item $\dagger$: R is ReLU and S is Swish.
      \item $\diamond$: F-IRB indicates Fused-Inverse-Residual-Block (Fused-IRB).
      \item $\ddagger$: E indicates the range of IRB expansion rate.
      \item ${*}$: number of filters increase from 24 to 32 at the step of 8. 
    \end{tablenotes}
  \end{threeparttable}
  \caption{The proposed convnet search space.}
  \label{table:search_space}
  \vspace{-.3cm}
 \end{table}

Table~\ref{table:search_space} demonstrates the details of our search space used in this paper. Our search space consists of 8 stages. Here we search for the configurations of each stage, and the layers within a stage share the same configurations. The first two stages are to search for the head configurations using convolutions. Inspired by EfficientNet-V2~\cite{tan2021efficientnetv2}, the 2 and 3 stages uses Fused-IRB~\cite{tan2021efficientnetv2}. But we observed the increasing latency after replacing the rest IRB with Fused-IRB. From the stage 4 to 7, we use IRB as the basic layers. The column \textit{\#Layers} shows the range of \#layers in the stage, e.g., [3, 10] at stage 4 means that the stage can have 3 to 10 IRBs. And the column \textit{Filters} shows the range of filters for the layers in the stage (see table note for details). Our search space also tunes the expansion ratio, activation types, kernel sizes, and the Squeeze Excitation(SE)~\cite{hu2018squeeze} layer inside the IRB/Fused-IRB. Finally, the dimensions of the input image increase from 224 to 512 at the step of 32.

\subsubsection{Justifications of the Search Space}
Unlike prior works, our search is guided by the accuracy and the TensorRT optimized inference latency. In a good experiment design, we should identify the most relevant factors~\cite{kempthorne1952design} to the design targets, i.e., fast and accurate networks. Table.~\ref{table:search_space} demonstrates the several prominent factors found by us to impact the latency and accuracy. Here we provide the empirical data to support our decisions.

\begin{figure}[t]
\vspace{-0.5cm}
\centering 
  \begin{center}
    \includegraphics[width=0.95\columnwidth]{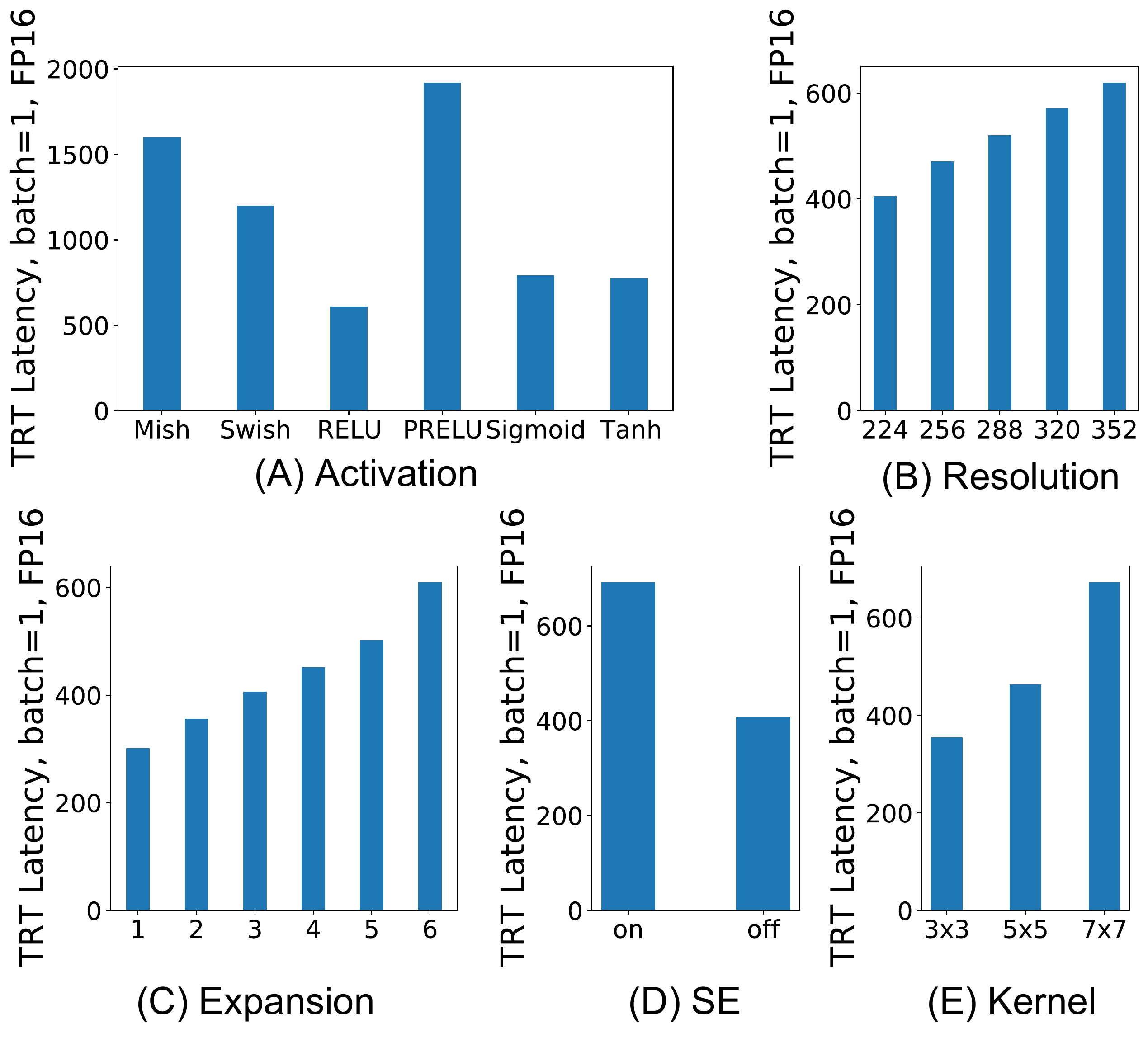}
  \end{center}
  \vspace{-0.3cm}
    \caption{The latency impact of different hyper-parameters and configurations on EfficentNet-B0.}
    \label{fig:search_space_justification}
\end{figure}

\begin{itemize}
  \item \#\textit{Layers} and \textit{Filters}: extensive evidence from published results demonstrates that a deep or wide network can perform better than the shallow or narrow variants~\cite{he2016deep, lu2017expressive} while adding layers or increasing filters slows down the inference. \#\textit{Layers} and \textit{Filters} are essential design choices to the accuracy and latency.
  \item \textit{Activation}: many past works have demonstrated that a good activation design can notably improve the final accuracy on ImageNet~\cite{he2015delving, ramachandran2017searching}, whereas Fig.~\ref{fig:search_space_justification}.A shows a network using ReLU can be 4x faster than using PReLU. In general, the activation is the memory-bound operation. And TensorRT supports ReLU, Sigmoid, and Tanh in the layer fusion, which explains the speed gap. Therefore, the choice of activation is an important factor in the latency and accuracy trade-off.
  \item \textit{Expansion}: IRB or Fused-IRB internally expands the channel size using a 1x1 convolution, and the expansion ratio controls the size of the internal channel, i.e., expansion ratio x the input channel. The MobileNet \cite{sandler2018mobilenetv2} paper claims that the larger channel expansion will help improve the capacity of the network and expressiveness. Our empirical results are also consistent with the claim. For example, the accuracy of a network drops 4 points on ImageNet top-1 after reducing the expansion from 6 to 2, whereas increasing the expansion incurs non-negligible costs(Fig~\ref{fig:search_space_justification}.C). These data suggest the expansion ratio is an important factor to search.
  \item \textit{Kernel}: a large convolution kernel can increase the receipt field to improve the accuracy (more details in \cite{araujo2019computing}). Still, it also increases the latency (Fig.~\ref{fig:search_space_justification}.D), which validates the choice of kernel size into the search space. 
  \item \textit{SE}: Squeeze-Excitation~\cite{hu2018squeeze} was introduced by the winning entry to ILSVRC 2017 that improved 25\% over the previous year. After adding SE, Fig.~\ref{fig:search_space_justification} shows the latency significantly increases. This justifies SE to be a factor in the search space.
  \item \textit{Image Resolution}: EfficientNet~\cite{tan2019efficientnet} clearly demonstrates the accuracy improvement by increasing the resolution, and Fig.~\ref{fig:search_space_justification}.B shows the latency also significantly increases. So we search for the input image resolution for better accuracy and latency trade-off.
\end{itemize}

\begin{table}[!tb]
\footnotesize
\setlength{\tabcolsep}{0.2em}
  \centering
    \vspace{-0.5cm}
    \begin{tabular}{ l l l l}
    \toprule
          {Stage} & {Type} &  Hyper-parameters & Length\\
    	  \midrule
    	          & Resolution & [\textit{Resolution}] & 1 \\
		  0       & Conv       & [\textit{\#Filters}]     & 1 \\
		  1       & Conv       & [\textit{Kernel, Activation, \#Layers}] & 3 \\
		  2       & Fused-IRB  & [\textit{\#Filters, Kernel, E, SE, Act, \#Layers}] & 6 \\
		  3       & Fused-IRB  & [\textit{\#Filters, Kernel, E, SE, Act, \#Layers}] & 6 \\
		  4       & IRB        & [\textit{\#Filters, Kernel, E, SE, Act, \#Layers}] & 6 \\
		  5       & IRB        & [\textit{\#Filters, Kernel, E, SE, Act, \#Layers}] & 6 \\
		  6       & IRB        & [\textit{\#Filters, Kernel, E, SE, Act, \#Layers}] & 6 \\
		  7       & IRB        & [\textit{\#Filters, Kernel, E, SE, Act, \#Layers}] & 6 \\
	\midrule
            &     & total      & 41 \\
    \bottomrule
    \end{tabular}
    \vspace{-0.2cm}
     \caption{The encoding scheme of networks in the search space.}
     \label{table:encoding_scheme}
     \vspace{-0.2cm}
\end{table}

\subsubsection{Network and Search Space Representations}
Now we have the general picture of search space; the next is to find the proper representation that embodies the design. We use a vector of integers to encode a network sampled from the search space described in Table.~\ref{table:search_space}. The length of the vector is 41, and Table.~\ref{table:encoding_scheme} elaborates the hyper-parameters represented by each digit. Stage 1 shares the same filter number as stage 0, and we only search the filter size for the first 3x3 convolution (stage 0). Because layers within a stage share the configurations, we use 6 integers to represent the filter size, kernel size, expansion ratio, using SE or not, the types of activation, and the number of IRBs for the stages from 2 to 7. So a network is an instance of the vector described in Table.~\ref{table:encoding_scheme}, and the range of every digit collectively define the search space in Table.~\ref{table:search_space}.

\begin{figure}[t]
\centering 
\vspace{-0.5cm}
  \begin{center}
    \includegraphics[width=0.6\columnwidth]{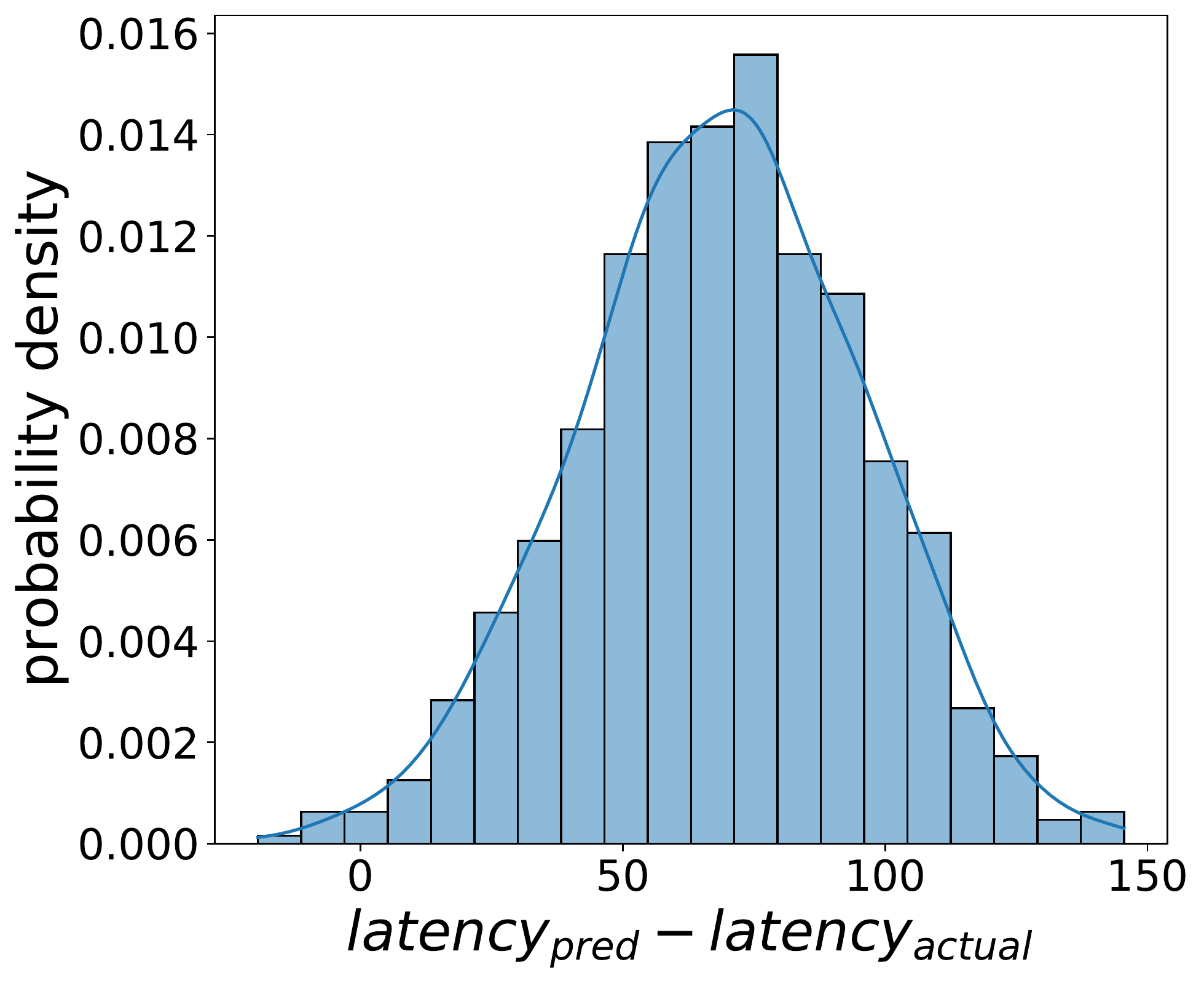}
  \end{center}
    \caption{The error between the actual measured TensorRT latency and the table predicted latency.}
    \label{fig:table_actual_latency}
    \vspace{-0.5cm}
\end{figure}

\subsection{Stratify Networks by Inference Latency}
\label{sec:gen_search_space}

\begin{figure}[t]
\centering 
\vspace{-0.5cm}
  \begin{center}
    \includegraphics[width=0.8\columnwidth]{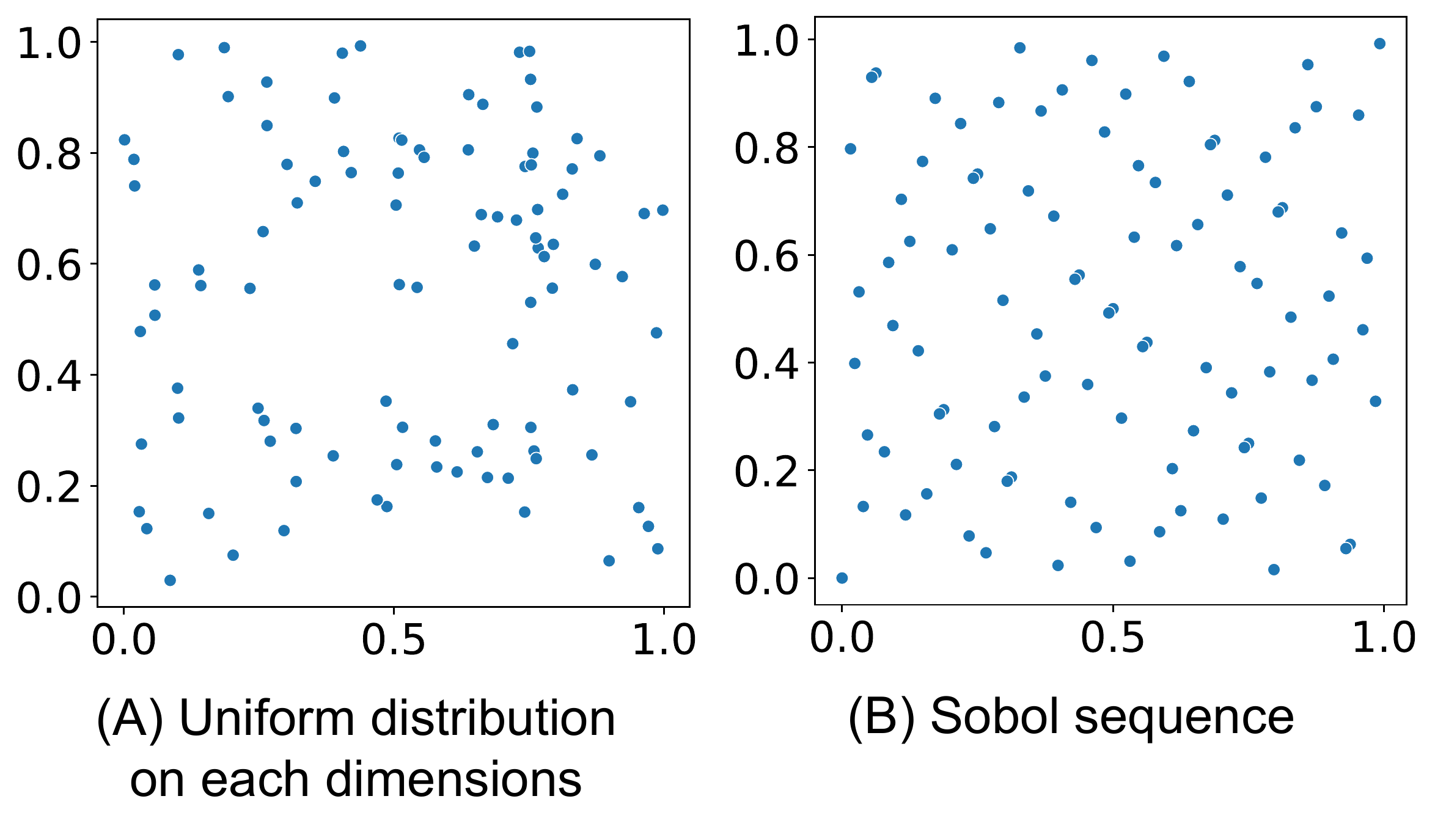}
  \end{center}
    \vspace{-0.5cm}
    \caption{The distribution sampled by (A) the uniform random on each dimensions and (B) sobol sequence.}
    \label{fig:sobolvsrandom}
\end{figure}

To design networks tiered by the inference latency, we choose to directly measure the latency of networks in the search space. Because the size of search space is exponentially large, we approximate the search space by sampling millions of networks from it. The sampling techniques is critical to capture the true distribution of search space, and here we use the Sobol sequence~\cite{sobolseq}, the advantages of which is straightforward in Fig.~\ref{fig:sobolvsrandom}. The sampling is a low cost operations that we can get millions of samples within a minute. The challenge is to measure the latency of sampled networks. Since TensorRT has dominated the MLPerf inference benchmark, we want to measure the inference latency optimized by TensorRT. Whereas, TensorRT takes minutes to build the inference engine for the measurement, which makes it infeasible to measure all the sampled networks.

We approximate a network's latency by adding up the latency of each layer. Although the search space renders $10^{30}$ networks, the layers have limited configurations, e.g., $10^4$ in our case. Therefore we can significantly speed up the latency measurement by building a latency table with the input data shape and the layer configurations as the key. Given a network, we iterate over layers to look up the latency. If a layer does not exist in the table, we only benchmark it and record its latency in the table. Finally, the network latency is the sum of the latency of all the layers. Fig.~\ref{fig:table_actual_latency} demonstrates that the table estimated latency is close to the network's actual latency, and the table estimation is on average 75$\mu s$ higher than the actual end-to-end measurement. Because a whole network subjects more opportunities for the layer fusion to TensorRT than the single layer. Benchmarking $\sim10^4$ layers is still an expensive task, and we parallelize the curation of the latency table over multi-GPUs to speed up the process from weeks to days.

\subsection{Distributed Neural Architecture Search}
We treat the network design as a black box. Guided by the reward, e.g. the validation accuracy, the search tunes the hyper-parameters prescribed in the search space to optimize the model performance~(Fig.\ref{fig:nas_framework}.A). Here we elaborate the details of search algorithms and the evaluation method.

\subsubsection{Search Algorithms}
We choose LA-MCTS boosted Bayesian optimization (BO)~\cite{wang2020learning} as the search algorithm, which is one of the top entries to the 2020 NeurIPS black-box optimization competition~\cite{BBOC}. Since we evaluate a network by training, the sample efficiency is critical to the overall cost. The competition results show that LA-MCTS boosted BO has demonstrated the leading sample efficiency among other BO variants and evolutionary algorithms; therefore, we adopt it in our experiments, and Fig.~\ref{fig:nas_framework} depicts the workflow.

Some prior works~\cite{lu2019nsga} define the problem as a Multi-Objective Optimization (MOO) to find the optimal Pareto frontier to the latency and accuracy. However, finding the Pareto Optimality is too fine-grained to the practice. For example, two solutions located on the Pareto frontier may have trivial differences in accuracy and latency, but finding these Pareto solutions is very expensive. Therefore, we organize the search space by latency before maximizing the accuracy. This also allows us to build a table of networks tiered by their inference latency.

\subsubsection{Evaluation}
\label{sec:evaluation}
We choose to evaluate each proposed network by training and apply the early-stopping if the training curve is not promising. For networks from the same search space with similar latency, we use the same training receipt, as our practical experience suggests that tuning the training receipt brings up to $1\%$ accuracy improvement at a tremendous cost. The details of our training receipts can be found at sec~\ref{training_receipts} in the supplemental material. The training will return the best validation accuracy to the search algorithm after 450 epochs.

Rather than democratizing NAS, this paper intends to maintain a set of NAS-optimized models using hundreds of GPUs for the community. We believe the training approach is necessary, although it is far more expensive than recent supernet approaches~\cite{pham2018efficient}. First, extensive evidence in \cite{zhao2021few, yu2019evaluating} demonstrates that supernet can be inaccurate in ranking the network candidates, and training a good supernet is non-trivial~\cite{yu2020train}. Second, no supernet properly supports variable expansion ratios, image resolutions, and with/without SE. The training approach can circumvent all these problems at additional costs.

\subsubsection{Distributed NAS}
Now we're ready to put everything together. Fig.~\ref{fig:nas_framework} demonstrates that we implement a client and server distributed system to run NAS. Following sec~\ref{sec:gen_search_space}, we start with generating networks in a latency range as the search space by sampling. Then we integrate the pruned search space into the search algorithm to run on the server. The server and clients exchange data via sockets. The clients will request a network from the server to evaluate if they are free and return the network and the best validation accuracy to the server. The search algorithm can leverage this information to propose the next network candidate. To validate the framework, we test the system with a few synthetic functions to ensure the performance metric increases along with the \#samples. This framework is also generic to different search problems, and we can also use the same framework to search the architecture for Transformer.

\section{Experiments}

This section demonstrates the details of using the proposed search space and NAS system in designing GPUNet. Compared to existing works, GPUNet significantly improves the SOTA Pareto frontier in both the accuracy and inference latency~(Fig.~\ref{teaser_figure}). With a similar latency of 1.8ms, the accuracy of GPUNet is 1\% better than the corresponding FBNet-V3 on ImageNet. With a similar 80.5 accuracy, GPUNet is 1.6x faster than FBNet. We start with describing the experiment setup, then discuss the main results. Finally, we show that GPUNet also effectively improves the downstream tasks.

\subsection{Experiment Setup}
\textbf{Software Setup}: we perform NAS directly on ImageNet~\cite{deng2009imagenet} that contains 1.28 million training and 50000 validation images in 1000 classes. Each network candidate is pre-trained with 300 epochs for the performance ranking with automatic mixed precision (AMP), then we fine-tune the top network for another 150 epochs. We use a modified training script from Pytorch Image Models~\cite{pytorchimagemodels} to train models. The training only uses random augmentation at the magnitude of 9 and a standard deviation of 0.5. The learning rate decays by .97 for every 2.4 epochs. Exponential Moving Average (EMA) is also in use with a decay rate of 0.9999. The crop percentage is set to 1, and the optimizer is RMSprop. We set NAS to focus on models with \texttt{FP32+FP16} TensorRT GPU compute time\footnote{TensorRT also reports throughput, end-to-end, device-to-host, host-to-device time. We use GPU compute time to better capture the latency impact on architecture difference.} $<2ms$, which is more relevant in practice. For latency measurement, we use TensorRT-8.0.1. We export the onnx model and measure FP16 GPU compute latency using the \texttt{trtexec --fp16} command-line on a standalone PCI-E NVIDIA GV100 GPU.

\textbf{Machine Setup}: we perform the training on DGX A100 with 8x A100 80 GB. Our system is flexible to allow training on preemptible (spot) instances. We launch the server at a dedicated node to propose networks and launch clients at preemptible instances. Each client checkpoints per epoch during the training in case of preemption. The server also consistently checkpoints its state for fault tolerance.

\subsection{Main Results}

\begin{table*}[!tb]
\vspace{-0.8cm}
\centering
\begin{threeparttable}
    \small	
    \centering
    \begin{tabular}{l l l l l l l}
          \multicolumn{7}{c}{Without Distillation} \\
          \midrule
          {}     & Top1     & TensorRT Latency$^{\dagger}$ &  \#Params & \#{\tt FLOPS} & GPUNet & GPUNet \\
          Models & ImageNet & FP16 GV100 (ms)          & (Million) & (Billion)    & Speedup $\uparrow$ & Accuracy $\uparrow$\\
          \midrule                                     
          RegNet-X~\cite{radosavovic2020designing}    & 77.0 & 2.06 & 9.19. & 1.6  & 3.3x   & 1.9 \\
          EfficientNet-B0~\cite{tan2019efficientnet}  & 77.1 & 1.18 & 5.28   & 0.38 & 1.9x  & 1.8 \\
          EfficientNetX-B0-GPU~\cite{li2021searching} & 77.3 & 1.05 & 7.6    & 0.91 & 1.69x & 1.6 \\ 
          FBNetV2-L1~\cite{wan2020fbnetv2}            & 77.2 & 1.13 & x      & 0.32 & 1.82x & 1.7 \\
          \textbf{GPUNet-0} & \textbf{78.9} & \textbf{0.62} & 11.9  & 3.25   &      \\
    	  \midrule                                     
    	  RegNet-X                                    & 80.0 & 3.9. & 54.3. & 15.  & 2.72x & 0.5\\
		  EfficientNet-B2                             & 80.3 & 1.86 & 9.2   & 1    & 2.26x & 0.2\\
		  EfficientNetX-B2-GPU                        & 80.0 & 1.61 & 10    & 2.3  & 1.96x & 0.5\\
		  FBNetV3-B~\cite{dai2021fbnetv3}             & 79.8 & 1.55 & 8.6.  & 0.46 & 1.89x & 0.7\\
		  ResNet-50~\cite{he2016deep}                 & 80.3 & 1.1  & 28.09 & 4.   & 1.34x & 0.2\\
		  \textbf{GPUNet-1} & \textbf{80.5} & \textbf{0.82}  & {12.7} & {3.3} & ~    \\ 
		  \midrule
		  RegNet-X                                    & 80.5 & 5.7  & 107.  & 31.7   & 3.24x & 1.7\\
		  EfficientNet-B3                             & 81.6 & 2.3  & 12    & 1.8    & 1.3x  & 0.6\\
		  EfficientNetX-B3-GPU                        & 81.4 & 2.1  & 13.3  & 4.3    & 1.2x  & 0.8\\
		  ResNeSt-50~\cite{zhang2020resnest}          & 81.1 & 2.3  & 27.5  & 5.4    & 1.27x & 1.1\\
 		  FBNetV3-F                                   & 82.1 & 2.26 & 13.9  & 1.18   & 1.28x & 0.1\\
           \textbf{GPUNet-2} & \textbf{82.2} & \textbf{1.76}  &25.8   & 8.38   &  \\
		  \midrule
		  \multicolumn{7}{c}{With Distillation} \\
		  \midrule
		  AlphaNet-a2~\cite{wang2021alphanet}         & 79.2 & 1.14  & 11.3  & 0.32   & 1.8x & 0.5\\
		  FBNetV3-A                                   & 79.6 & 1.52  & 8.6   & 0.35   & 2.4x & 0.1\\
		  BigNAS-L~\cite{yu2020bignas}                & 79.5 & 1.55  & 6.4   & 0.58   & 2.46 & 0.2\\
		  LaNet-200M                                  & 77.8 & 0.95  & 5.76  & 0.24   & 1.5x & 1.9\\
		  \textbf{GPUNet-D0} & \textbf{79.7} & \textbf{0.63} &   6.2 & 0.72 &      \\
		  \midrule
		  AlphaNet-a6                                 & 80.8 & 1.6  & 15.4  & 0.71   & 1.28x & 1.7\\
		  FBNetV3-D                                   & 81.1 & 1.7  & 10.3  & 0.64   & 1.36x & 1.4\\
		  BigNAS-XL                                   & 80.9 & 1.64 & 9.5   & 1.04   & 1.31x  & 1.6\\
		  LaNet-600M                                  & 80.8 & 1.29 & 8.67  & 0.24  & 1.03x  & 1.7\\
		  \textbf{GPUNet-D1} & \textbf{82.5} & \textbf{1.25} &  10.6 & 3.66 &      \\
		  \midrule
		  FBNetV3-G                                   & 83.2 & 2.43 & 10.8   & 2.1   & 1.0x  & 0.4\\
		  \textbf{GPUNet-D3} & \textbf{83.6} & \textbf{2.4}  &  19 & 15.6    & \multicolumn{2}{c}{(Only random augmentation)} \\
		  \bottomrule
		  \label{table:nas_results}
    \end{tabular}
    \vspace{-0.15cm}
    \begin{tablenotes}
      \item $^\dagger$: We measure the latency (\texttt{FP16} GPU compute time) using an explicit-shape at batch size 1.
    \end{tablenotes}
    \vspace{-0.2cm}
    \caption{Comparisons of GPUNet to SOTA results. Fig.~\ref{teaser_figure} visualizes the table and shows that GPUNet-D dominates the baseline models in both the accuracy and inference latency.}
    \label{table:nas-results}
\end{threeparttable}
\end{table*}

\begin{figure}[t]
\centering 
  \begin{center}
    \includegraphics[width=0.8\columnwidth]{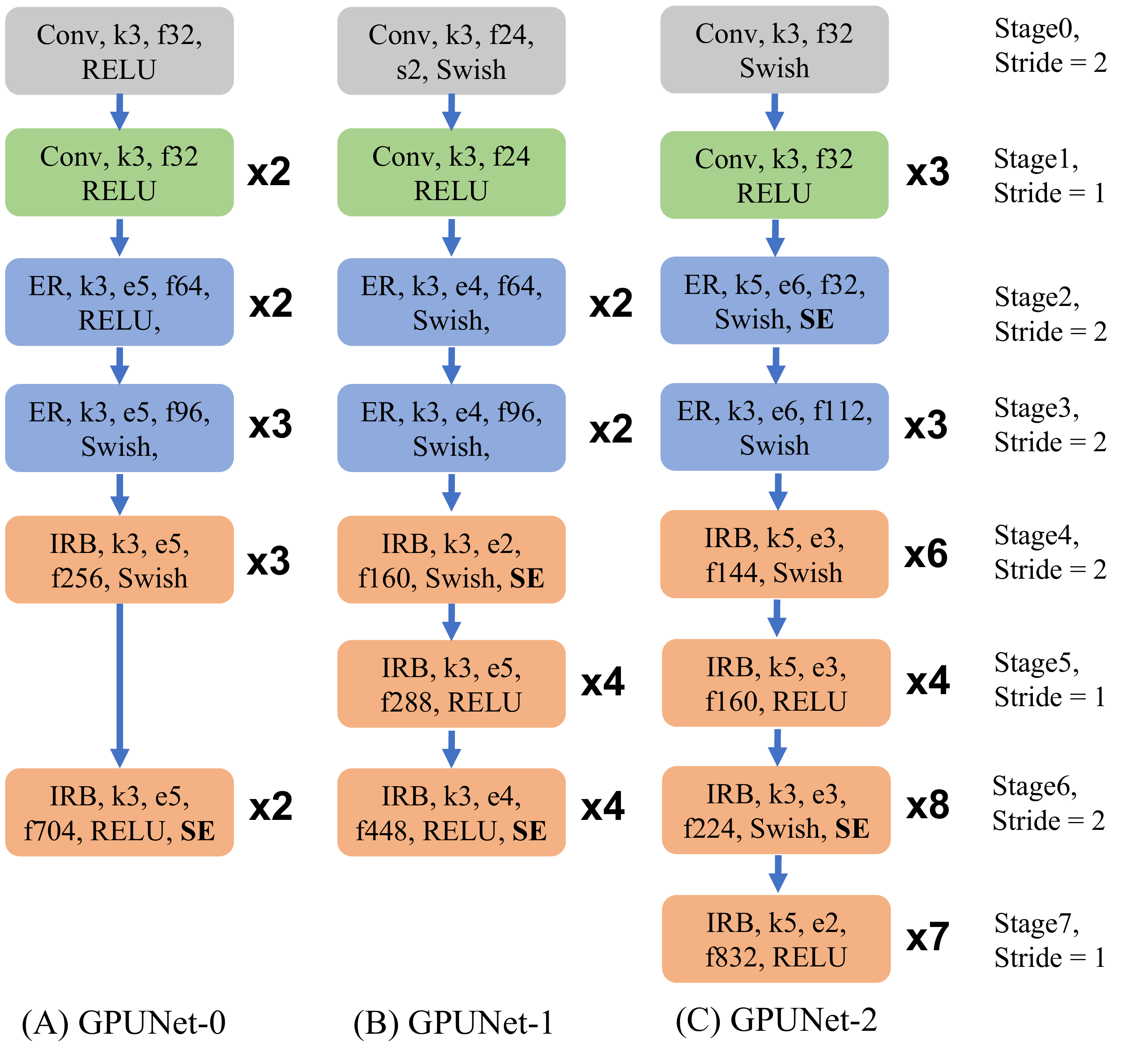}
  \end{center}
    \vspace{-0.5cm}
    \caption{The architecture of searched GPUNet in Table.~\ref{table:nas-results}. For stages 2, 3, 4 and 6, only the stride of first layer is 2 and the stride of rest layers is 1.}
    \label{fig:gpunet_architecture}
\end{figure}

\subsubsection{Preparation of Baselines}
Table.~\ref{table:nas-results} lists the recent SOTA baselines used in comparisons; There are two set of baselines that include and exclude distillation, respectively. The accuracy and the inference latency are two key metrics in our evaluations. Comparing the accuracy is easy to be fair but not on the inference latency, since the latency can be impacted by the software stack (e.g. runtime efficiency and system optimizations), GPUs, batch size and etc.. To ensure fairness, we transform all the baseline models into ONNX models for benchmarking with TensorRT. The inference benchmark exclusively runs on an NVIDIA GV100; and the workspace of TensorRT is fixed at 10G across all runs. We benchmark the latency at the batch size = 1 with a explicit-shape, and report the average latency from 1000 runs. We take baseline models from either their original implementations or from PyTorch Image Models~\cite{pytorchimagemodels}. Appendix Sec.~\ref{baseline_sources} provides the details.

\subsubsection{Results on ImageNet1K}
Table.~\ref{table:nas-results} compares the performance of searched GPUNet to baselines, and Fig.~\ref{teaser_figure} visualizes the Pareto frontier of different models in the inference latency and accuracy. Please note we prepare two sets of different networks GPUNet and GPUNet-D, for the cases of with/without distillation. Fig.~\ref{teaser_figure}.B clearly shows GPUNet-D dominates other models in both target objectives, and Table.~\ref{table:nas-results} suggests both GPUNet and GPUNet-D are significantly faster than SOTA networks while maintaining similar accuracy. At the similar 80.5 top-1 accuracy, GPUNet-1 is nearly 2 times faster than EfficientNet-B2, EfficientNetX-B2-GPU, and FBNetV3-B. For other accuracy groups, GPUNet consistently demonstrates the speedup from $1.27\times$ to $3.24\times$ than baselines. 

While EfficientNet-V2~\cite{tan2021efficientnetv2} shows slightly better results than GPUNet when latency $> 3ms$ in Fig.~\ref{teaser_figure}.B, EfficientNet-V2 utilizes a far more sophisticated training scheme that includes Mixup~\cite{zhang2017mixup} and progressive training in regularizing the network for better accuracy. These regularizations are orthogonal to NAS, and they can be an excellent future work to improve the accuracy of GPUNet further. 

Interestingly, we also note that the \#{\tt FLOPS} and \#Parameters of GPUNet are larger than baselines, though GPUNet is significantly faster. These results indicate that low {\tt FLOPS} models are not necessarily fast on GPUs. EfficientNet-X explains this with the roofline model~\cite{li2021searching}, and we will provide more results in Sec.~\ref{sec:gpunet_explaination}.

Please note that our ultimate goal is to provide a table of models tiered by their inference latency to expedite the customization. Table.~\ref{table:nas-results} and Fig.~\ref{teaser_figure} only show a few models to demonstrate that our NAS system can effectively design fast and accurate networks on the proposed search space in Table.~\ref{table:search_space}. We will release this table of models after the paper.

\subsubsection{GPUNet Architecture}
\label{sec:fig:gpunet_architecture}
Fig.~\ref{fig:gpunet_architecture} shows that the architectures of NAS optimized GPUNet are too irregular to be human design. For example, the two adjacent stride=2 ER (Fused-IRB) blocks in GPUNet-2 consecutively halve H and W twice, while the human-designed networks usually have multiple stride = 1 layer between two stride = 2 layers. There is no obvious pattern for the activation functions and expansions in IRB as well. However, these NAS optimized networks show one common characteristic in the filter distribution, which are skinny in the beginning/middle stages and very wide in the last few stages though the search space in Table.~\ref{table:search_space} permits large filters at the beginning and small filters in the end. For example, the filters of GPUNet-2 follow the pattern of $32\rightarrow32\rightarrow116\rightarrow144\rightarrow160\rightarrow224\rightarrow832$; GPUNet-0 and GPUNet-1 also follow a similar filter pattern.

\subsubsection{Why GPUNet Are Faster and Better?}
\label{sec:gpunet_explaination}
We also compare the architecture of FBNet and EfficientNet to GPUNet. Here are a few key differences found by us that explain the GPUNet performance. Let's use GPUNet-1 as an example.
\begin{itemize}
  \item \textit{Mixed types of activation}: Fig.\ref{fig:gpunet_architecture} suggests that GPUNet switches between RELU and Swish, but EfficientNet and FBNet use Swish across all the layers. Fig.\ref{fig:search_space_justification}.A suggests Swish greatly increases the latency. Some layers of GPUNet uses RELU to reduce the latency for other opportunities to improve the accuracy, e.g., larger filters.
  \item \textit{Fewer expansions in IRB}: Fig.\ref{fig:search_space_justification}.C shows the network latency almost doubles by increasing the expansions in all IRB from 1 to 6. The expansion is part of our search space, so some GPUNet layers tend to have small expansions to save the latency.
  \item \textit{Wider and Deeper}: the filters (wide) and the number of layers (deep) in a stage are part of our search space. Because of the latency saving from mixed activation and fewer expansions, GPUNet tends to be wider and deeper than FBNet and EfficientNet. In the same accuracy group, the filters of FBNetV3-B follow the pattern of $16\rightarrow24\rightarrow40\rightarrow72\rightarrow120\rightarrow183\rightarrow224$, and the filter pattern of EfficientNet-B2 is $32\rightarrow16\rightarrow24\rightarrow48\rightarrow88\rightarrow120\rightarrow208\rightarrow352$, but GPUNet-1 is a lot wider than FBNetV3-B and EfficientNet-B2 that has a pattern of $24\rightarrow64\rightarrow96\rightarrow160\rightarrow288\rightarrow448$. Besides, GPUNet-2 has 33 layers, 2 more than FBNetV3-F and 5 more than EfficientNet-B3. It is known that deep and wide networks have better accuracy; therefore, the accuracy of GPUNet is better than baselines within each group.
  \item \textit{Larger Resolution}: GPUNet-(1 and 2) are 32 and 64 larger than EfficientNet-B2 and B3 in resolutions, 72 and 120 larger than FBNetV3-B and FBNetV3-F, respectively. Using large resolution generally improves the accuracy; therefore, GPUNet shows better accuracy and higher {\tt FLOPS} than baselines.
\end{itemize}

\begin{table}[!tb]
\footnotesize
\setlength{\tabcolsep}{0.2em}
  \centering
    \begin{tabular}{l l l l l }
          \toprule
          {Backbone}    & ImageNet top1 & {Method} & TRT Latency(ms) & mAP \\
          \midrule
          GPUNet-2      & 82.2 &  Cascade RCNN & 5.2 & 40.0      \\
          ResNet-50     & 80.3 &  Cascade RCNN & 5.8 & 40.4      \\
          FBNetV3-F     & 82.1 &  Cascade RCNN & 7.90 & 26.5     \\
          EfficientNet-B3 & 81.6 &  Cascade RCNN & 10.65 & 28.40 \\
          \bottomrule
    \end{tabular}
     \caption{Applying GPUNet to COCO object detection tasks. The latency was measured using the resolution of 1333x800.}
     \label{table:gpunet_coco}
\end{table}

\subsection{Evaluating on Detection tasks}
We test GPUNet on COCO detection tasks. We evaluate GPUNet, FBNetV3-F, and EfficientNet-B3 on COCO detection tasks by replacing the backbone in the cascade RCNN~\cite{cai2018cascade}. Table.~\ref{table:gpunet_coco} shows GPUNet-2 is not only faster, but also delivers higher mAP than baseline models.

\section{Conclusions}
Model customization is challenging for DL engineering. In this work, we proposes to build a model hub tiered by their inference latency to expedite model customization, which facilitates practitioners to reuse the our pretrained models with known TensorRT latency and compatibility in mind. With a novel distributed NAS system and an enhanced search space to design a set of fast and accurate GPUNet, we establish a new SOTA Pareto frontier in latency and accuracy, validating the effectiveness of our NAS system. Although this paper primarily focus on EfficientNet search space, our NAS system is generic to support various tasks and search spaces. Ultimately we intend to maintain a hub of NAS optimized models that track the latest technology so that ML practitioners can directly reuse them.

{\small
\bibliographystyle{ieee_fullname}
\bibliography{egbib}
}

\newpage
\section{Supplemental Material}
\subsection{Training Receipts}
\label{training_receipts}

\begin{table}[!tb]
\footnotesize
\setlength{\tabcolsep}{0.2em}
  \centering
    \begin{tabular}{l l l }
          \toprule
          {Hyper-parameters} & {Value}   \\
          \midrule
          sched              & step      \\
          decay-epochs       & 2.4       \\
          decay-rate         & 0.97      \\
          opt                & rmsproptf \\
          b                  & 192       \\
          epochs             & 450       \\
          opt-eps            & 0.001     \\
          j                  & 8         \\
          warmup-lr          & 1e-6      \\
          weight-decay       & 1e-5      \\
          drop               & 0.3       \\
          drop-connect       & 0.2       \\
          model-ema          & True      \\
          model-ema-decay    & 0.9999    \\
          aa                 & rand-m9-mstd0.5 \\
          remode             & pixel     \\
          reprob             & 0.2       \\
          lr                 & 0.06      \\
          amp                & True      \\
          crop-pct           & 1.0       \\
          \bottomrule
    \end{tabular}
     \caption{The training hyper-parameters: we use Pytorch Image Models to train GPUNet, and here~\cite{pytorchimagemodels} further explains the usage of these hyper-parameters.}
     \label{table:training-hyperparameters}
\end{table}

Table.~\ref{table:training-hyperparameters} shows the full details of training hyper-parameters. We used Pytorch Image Models in training, and we applied the same configurations to all GPUNet. 

\subsubsection{Sources of Baseline}
\label{baseline_sources}
The baseline models are from their original public release to ensure fair evaluations. We only convert their models to ONNX so that we can benchmark them in TensorRT. The conversion is invasive to the model latency and structure, and we use the Pytorch and Tensorflow native support for ONNX conversions. Here is the list that shows the source of the original implementation.
\begin{itemize}
\item FBNet:\url{https://github.com/facebookresearch/mobile-vision}
\item EfficietNet-X:\url{https://github.com/tensorflow/tpu/blob/master/models/official/efficientnet/tpu/efficientnet_x_builder.py}
\item EfficientNet:\url{https://github.com/rwightman/pytorch-image-models}
\item RegNet:\url{https://github.com/facebookresearch/pycls}
\item AlphaNet:\url{https://github.com/facebookresearch/AlphaNet}
\item ResNeSt:\url{https://github.com/zhanghang1989/ResNeSt}
\item LaNet:\url{https://github.com/facebookresearch/LaMCTS}
\item OFA:\url{https://github.com/mit-han-lab/once-for-all}
\end{itemize}

\subsubsection{Verify the models on more devices}

\begin{figure}[t]
\centering
  \begin{center}
    \includegraphics[width=\columnwidth]{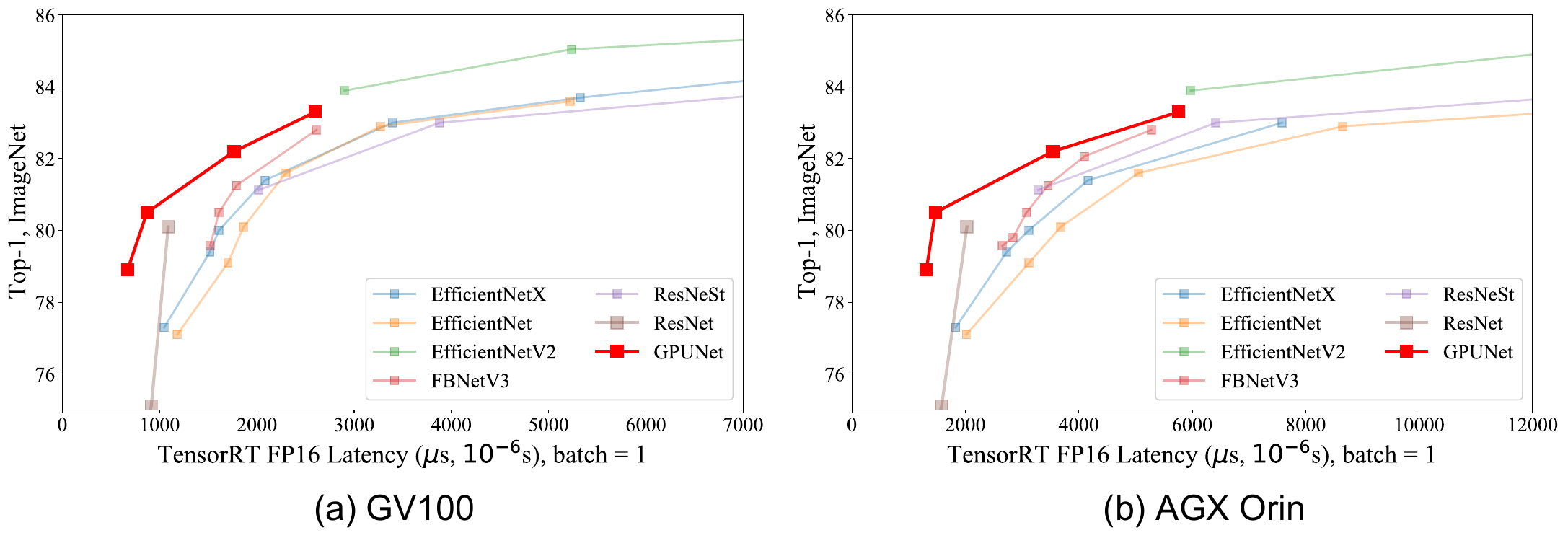}
  \end{center}
    \caption{GPUNet performance on AGX Orin and GV100.}
    \label{gpunet_orin}
\end{figure}

In Fig.~\ref{gpunet_orin}, we have tested GPUNet optimized for GV100 on NVIDIA AGX Orin, and GPUNet consistently dominates other networks in the accuracy and latency Pareto frontier. GPUNet maintains the advantages because it replaces some memory-bound operators (e.g., high expansion ratio in SE layers) to compute bound operators, such as larger filters or deeper networks. An interesting observation is that the advantages of GPUNet-2 and GPUNet-3 (latency $>$ 3000 on Orin) decrease on Orin w.r.t on GV100. Because GV100 has more execution units than Orin, increasing filters or layers can better saturate the device. Therefore, the optimization strategies generalized from GPUNet in sec.4.2.4 are still applicable to other devices.

\end{document}